\documentclass[conference]{IEEEtran}
\IEEEoverridecommandlockouts
\usepackage[ruled,vlined,linesnumbered]{algorithm2e}
\usepackage[normalem]{ulem}
\usepackage{cite}       
\usepackage{hyperref}   
\usepackage{amsmath}    
\usepackage{amssymb}    
\usepackage{amsfonts}   
\usepackage{bm}         
\usepackage{mathtools}  
\usepackage{enumitem}   
\usepackage{graphicx}   
\usepackage{multirow}   
\usepackage{siunitx}    
\usepackage{pifont}     
\usepackage{xcolor}     
\usepackage{eso-pic}    
\setlist{nosep}
\SetKwFor{ForAll}{for all}{do}{endfor}
\renewcommand{\arraystretch}{0.95}
\def\secref#1{Section~\ref{#1}}

\def\tabref#1{Table~\ref{#1}}
\def\eqref#1{(\ref{#1})}

\newcommand\etal{\emph{et~al.}}

\newcommand{\eg}{\emph{e.g.}}
\newcommand{\wrt}{w.r.t.\ }

\def\argmin{\mathop{\rm argmin}}
\def\atan2{\mathop{\rm atan2}}
\newcommand{\cmark}{\ding{51}}
\newcommand{\xmark}{\ding{55}}
\NewDocumentCommand{\B}{}{\bfseries}
\NewExpandableDocumentCommand{\U}{m}{#1\pU{}{#1}}
\NewExpandableDocumentCommand{\UB}{m}{\B#1\pU{\B}{#1}}
\NewDocumentCommand{\pU}{mm}{%
  \llap{\uline{\phantom{#1\num{#2}}}}%
}
\def\finalversion{}
\title{%
Unsigned Distance Maps on\\
2D Point Cloud Registration}
\ifdefined\finalversion%
\author{%
Ricardo B. Sousa$^{1,2}$,
Giorgio Grisetti$^{3}$,
H\'{e}ber Miguel Sobreira$^{2}$,\\
Carlos André Silva$^{4}$, and
Ant\'{o}nio Paulo Moreira$^{1,2}$%
\thanks{$^{1}$Faculty of Engineering, University of Porto (FEUP).
        R. Dr. Roberto Frias,
        4200-465 Porto, Portugal.
        \texttt{\{rbs,amoreira\}@fe.up.pt}}%
\thanks{$^{2}$INESC TEC -- Institute for Systems and Computer Engineering,
        Technology and Science.
        R. Dr. Roberto Frias,
        4200-465 Porto, Portugal.
        \texttt{heber.m.sobreira@inesctec.pt}}%
\thanks{$^{3}$Sapienza University of Rome.
        Piazzale Aldo Moro 5,
        00185 Roma RM, Italy.
        \texttt{grisetti@diag.uniroma1.it}}%
\thanks{$^{4}$Flowbotic Mobile Systems, S.A.
        Lugar do Pombal, Zona Industrial de Salgueiro,
        3530-259 Viseu, Portugal.
        \texttt{carlos.silva@flowbotic.eu}}%
\thanks{This work is co-financed by Component 5 -- Capitalization and Business
Innovation, integrated in the Resilience Dimension of the Recovery and
Resilience Plan within the scope of the Recovery and Resilience
Mechanism~(MRR) of the European Union~(EU), framed in the Next Generation~EU,
for the period 2021--2026, within project GreenAuto, with reference 54.}%
}%
\else%
\author{%
Removed for blind revision.%
\thanks{Removed for blind revision.}%
}%
\fi%
\begin{document}

\maketitle
\AddToShipoutPictureBG*{\AtPageLowerLeft{%
  \put(\LenToUnit{\dimexpr1in+\hoffset+\oddsidemargin\relax},
       \LenToUnit{\dimexpr\paperheight-0.6in\relax}){%
    \makebox[0pt][l]{\footnotesize\bfseries
      Preprint of the submitted version. Paper accepted at
      9th Iberian Robotics Conference (ROBOT2026).}}}}
\thispagestyle{empty}
\pagestyle{empty}


\begin{abstract}
2D~point cloud registration arises in laser odometry
and Simultaneous Localization and Mapping~(SLAM) for mobile robots.
Iterative Closest Point~(ICP) is one of the most widely used approaches.
Still, its iterative procedure recomputes correspondences via
nearest-neighbor search at every iteration, whereas correspondence-free
alternatives focus on scan-to-map alignment.
This paper proposes a 2D~point cloud registration approach based on
unsigned distance maps, precomputing the Euclidean distance to
the nearest reference point, along with its spatial derivatives,
over a discrete grid, replacing the per-iteration search with $O(1)$ lookups.
Moreover, point-to-point and point-to-plane error formulations are derived on
the $SE(2)$ manifold and solved via Gauss-Newton optimization.
On a synthetic benchmark and the real-world IILABS~3D dataset,
the precomputed point-to-point variant outperforms its analytical counterparts,
achieving competitive laser-odometry drift compared to
point-to-plane formulations, as the precomputed gradient regularizes
correspondences in the presence of sensor noise.
\end{abstract}



\section{Introduction}\label{sec:intro}

2D~point cloud registration estimates the rigid-body transformation
that best aligns two point clouds acquired by a range sensor.
This problem arises in laser odometry and
Simultaneous Localization and Mapping~(SLAM),
where scans from a 2D~laser are aligned to track
the robot's pose and build a consistent representation of the environment.
For ground mobile robots equipped with planar laser scanners, registration
typically serves as the front-end of a SLAM pipeline.

Iterative Closest Point~(ICP)~\cite{besl-mckay:tpami:1992,chen-medioni:icra:1991}
remains the most widely used registration approach, alternating between
data association, which establishes point correspondences via a nearest-neighbor
search (commonly accelerated with a kd-tree~\cite{bentley:cacm:1975} at
$O(\log n)$ cost per point), and an error-minimization step using
a point-to-point~\cite{besl-mckay:tpami:1992} or
point-to-plane~\cite{chen-medioni:icra:1991} metric. Methods such as
the Normal Distributions Transform~(NDT)~\cite{biber-strasser:iros:2003} and
probabilistic formulations~\cite{montesano:iros:2005,censi:icra:2006} avoid
explicit correspondences but are typically oriented toward scan-to-map alignment.
In contrast, precomputed spatial-gradient approaches, such as
HectorSLAM~\cite{kohlbrecher:ssr:2011} and
PerfectMatch~\cite{lauer:robocup:2006,sobreira:jint:2019}, evaluate the
registration error against occupancy grid maps or precomputed distance fields,
thereby removing per-iteration association.
Nonetheless, these approaches were oriented towards scan-to-map registration
under specific assumptions, leaving open the design of a unified representation
that removes the per-iteration nearest-neighbor search while supporting both
point-to-point and point-to-plane residuals within
a common least-squares framework.

This work proposes a 2D~point cloud registration approach based on unsigned
distance maps. A distance map encodes the Euclidean distance to the nearest
reference point, decoupling data association from the optimization loop.
During registration, each moving point is projected and queried in the distance
field at $O(1)$ cost, replacing the nearest-neighbor
search performed at every ICP iteration. We derive point-to-point and
point-to-plane error formulations on the $SE(2)$ manifold and solve them with a
Gauss-Newton optimizer within the srrg2\_solver
framework~\cite{grisetti:robotics:2020}. Our distance map data structure
supports three representation levels of increasing information and memory cost,
dense and sparse memory storage, while remaining compatible with the
ICP~\cite{besl-mckay:tpami:1992,chen-medioni:icra:1991} baselines.

The main contributions of this work are the following:
\begin{itemize}
\item a unified unsigned distance map open-source framework%
      \ifdefined\finalversion
      \footnote{\url{https://github.com/INESCTEC/ricoslam} \emph{(to be released)}}
      \else
      \footnote{Repository link removed for blind review.}
      \fi for 2D~point cloud registration, with three representation levels and
      supporting dense or sparse storage layouts, and serving as a drop-in
      replacement for the kd-tree~\cite{bentley:cacm:1975} data association of
      classical ICP~\cite{besl-mckay:tpami:1992,chen-medioni:icra:1991};
\item point-to-point and point-to-plane residuals and their Jacobians on the
      $SE(2)$ manifold built on the distance field, with the analytical
      point-to-plane proposed formulation shown to be equivalent to classical
      point-to-plane ICP~\cite{chen-medioni:icra:1991};
\item an experimental evaluation on a synthetic benchmark and the real-world
      IILABS~3D~\cite{ribeiro:access:2025} dataset, showing that the precomputed
      point-to-point variant outperforms its analytical counterparts and
      achieves competitive odometric drift compared to point-to-plane
      formulations, as the precomputed finite-difference gradient regularizes
      point-to-point correspondences against sensor noise.
\end{itemize}

The remainder of this paper is organized as follows.
\secref{sec:related} reviews literature on 2D~and 3D~point cloud registration.
\secref{sec:dmap} introduces the unsigned distance map representation and its
data structures. \secref{sec:method} formulates the proposed point-to-point and
point-to-plane registration on the $SE(2)$ manifold. \secref{sec:results}
presents the experimental evaluation on synthetic and real-world data. Lastly,
\secref{sec:conclusions} presents the conclusions and directions for future work.


\section{Related Work}\label{sec:related}

One of the most popular approaches for point cloud registration is
the ICP algorithm. Besl and McKay~\cite{besl-mckay:tpami:1992} introduced
ICP for 3D~shapes, iterating between (i)~computing
correspondences via nearest-neighbor and
(ii)~estimating the rigid transformation that minimizes a point-to-point
distance between corresponding points.
Chen and Medioni~\cite{chen-medioni:icra:1991} proposed a point-to-plane
variant exploiting surface normals on the target cloud,
minimizing the squared dot product of the normal with the residual between
projected source and target points.
Comparing ICP variants, Rusinkiewicz and
Levoy~\cite{rusinkiewicz-levoy:im:2001} found
point-to-plane~\cite{chen-medioni:icra:1991} to outperform the
point-to-point~\cite{besl-mckay:tpami:1992} metric.

Unlike ICP methods that rely on explicit correspondences, NDT by Biber and
Strasser~\cite{biber-strasser:iros:2003} performs 2D~registration by
discretizing space into cells, each modelled as a Gaussian over its points,
and minimizing an error function derived from these distributions via
Gauss-Newton.
Magnusson~\etal~\cite{magnusson:jfr:2007} extended
NDT~\cite{biber-strasser:iros:2003} to 3D~using voxels, aligning scans
by evaluating point likelihoods under the other scan's distributions.
Compared to point-to-point ICP~\cite{besl-mckay:tpami:1992} with kd-tree
search~\cite{bentley:cacm:1975}, 3D-NDT~\cite{magnusson:jfr:2007}
lowered computation cost and offered a more compact representation by avoiding
explicit correspondences.

While using
NDT~\cite{biber-strasser:iros:2003,magnusson:jfr:2007} representations,
GICP~\cite{segal:rss:2009} and
NICP~\cite{serafin-grisetti:iros:2015} retain the ICP iterative procedure.
GICP~\cite{segal:rss:2009} kept
kd-tree~\cite{bentley:cacm:1975} correspondence search but introduced a
plane-to-plane metric, modelling local surface covariances via
Principal Component Analysis (PCA)
to weight residuals by geometric uncertainty. GICP proved more robust and
accurate than point-to-point and point-to-plane
ICP~\cite{besl-mckay:tpami:1992,chen-medioni:icra:1991}.
NICP~\cite{serafin-grisetti:iros:2015} extended
GICP~\cite{segal:rss:2009} in both association (a line-of-sight step
with normal- and curvature-based rejection) and optimization, minimizing a
Mahalanobis distance in a 6D~point-normal space. NICP outperformed
GICP~\cite{segal:rss:2009} and NDT~\cite{magnusson:jfr:2007} in
accuracy.

Probabilistic methods have also been proposed for 2D~registration.
Montesano~\etal~\cite{montesano:iros:2005} build correspondences
from a Mahalanobis-based test and maximize the alignment
probability over the correspondence distribution, converging faster than
ICP~\cite{besl-mckay:tpami:1992}. Censi~\cite{censi:icra:2006} considered all
possible correspondences, generating alignment hypotheses through the
Generalized Hough Transform (GHT) within a probabilistic framework.
The GMapping~\cite{grisetti:tro:2007} and Cartographer~\cite{hess:icra:2016}
SLAM algorithms use Maximum Likelihood Estimation (MLE) formulations to align
scans against the map. GMapping~\cite{grisetti:tro:2007} performs a
gradient-descent search over the observation likelihood on the grid map using
the beam-endpoint model~\cite{thrun:book:2005}.
Cartographer~\cite{hess:icra:2016} formulates
scan-to-submap matching as a non-linear least-squares problem over smoothed
occupancy values.

Precomputed spatial gradients enable 2D~registration without explicit
correspondences. HectorSLAM~\cite{kohlbrecher:ssr:2011} maximizes the
occupancy probability of beam endpoints over the 2D~grid map, using
bilinear interpolation and approximate derivatives in a Gauss-Newton update.
PerfectMatch~\cite{lauer:robocup:2006,sobreira:jint:2019} formulates
2D~registration on distance maps for robot soccer, evaluating each point
through a distance map giving the distance to the
closest field marking under a robust M-estimator. This method
considered only scan-to-map alignment and approximated the estimation
covariance assuming orthogonal walls. Benchmarking against
ICP~\cite{besl-mckay:tpami:1992} and NDT~\cite{biber-strasser:iros:2003},
Sobreira~\etal~\cite{sobreira:jint:2019} found
PerfectMatch~\cite{lauer:robocup:2006} faster given the usage of precomputed
gradients and more tolerant to orientation initialization errors.

Overall, most ICP methods require two steps per iteration while probabilistic
approaches typically solve a Maximum A Posteriori~(MAP) problem and need a map
estimate over time, which is incompatible with scan-to-scan
registration~\cite{segal:rss:2009}.
In contrast, distance maps provide precomputed lookup tables that evaluate
each source point's registration error without explicit association,
supporting both scan-to-scan and scan-to-map alignment. Although
PerfectMatch~\cite{lauer:robocup:2006} targeted scan-to-map registration,
adapting it should enable scan-to-scan alignment, with faster computation
than ICP~\cite{besl-mckay:tpami:1992} and
NDT~\cite{biber-strasser:iros:2003}, as shown by
Sobreira~\etal~\cite{sobreira:jint:2019}.


\section{Unsigned Distance Maps}\label{sec:dmap}

Let $\mathcal{P} = \{\mathbf{p}_1, \dots, \mathbf{p}_n\} \subset \mathbb{R}^2$
denote a reference target 2D~point cloud acquired by a planar laser scanner and
expressed in the sensor frame $S$. The distance map
$\mathcal{D}: \mathbb{R}^2\to\mathbb{R}_0^+$ maps a query point
$\mathbf{y}\in\mathbb{R}^2$ represented in the sensor frame $S$
to its Euclidean distance to the nearest point in
$\mathcal{P}$, as in \eqref{eq:dmap:def}, where
$\mathbf{p}^*(\mathbf{y}) = \argmin_{\mathbf{p}_i \in \mathcal{P}}
\|\mathbf{y}-\mathbf{p}_i\|_2$ is the nearest neighbor.
\begin{equation}\label{eq:dmap:def}
\mathcal{D}(\mathbf{y}) =
\{\mathbb{R}^2\to\mathbb{R}_0^+ : \|\mathbf{y}-\mathbf{p}^*(\mathbf{y})\|_2\}
\end{equation}
Since $\mathcal{D}(\mathbf{y})\geq 0$ for all $\mathbf{y}$, it is an
\emph{unsigned distance function}, as opposed to signed distance functions
used in volumetric reconstruction. Unsigned distance fields are continuous
everywhere but not everywhere differentiable~\cite{jones:tvcg:2006}.

Whereas classical ICP~\cite{besl-mckay:tpami:1992,chen-medioni:icra:1991}
recomputes correspondences at every iteration because the pose estimate keeps
changing, we precompute the distance map once over a discrete grid of cell
resolution $\delta$ (m/px) and query its cell value without any further
reference to $\mathcal{P}$. Each cell stores information about the nearest
reference point, solving implicitly data association within the error evaluation
step: we transform a moving point by the current pose estimate and, at the
resulting cell, the lookup returns the distance and its derivatives in $O(1)$.

\subsection{Spatial derivatives}\label{sec:dmap:deriv}

Wherever the nearest neighbor is unique, $\mathcal{D}$ is smooth and its
gradient is the unit vector pointing from $\mathbf{p}^*(\mathbf{y})$ towards
$\mathbf{y}$, as in \eqref{eq:dmap:grad}. The unit-norm property
$\|\bm{\nabla}\mathcal{D}\|_2 = 1$ holds almost everywhere (Eikonal
property of Euclidean distance fields~\cite{jones:tvcg:2006}) except at points
equidistant from two or more reference points, where the gradient is undefined.
\begin{equation}\label{eq:dmap:grad}
\bm{\nabla}\mathcal{D}(\mathbf{y}) =
\frac{\mathbf{y}-\mathbf{p}^*(\mathbf{y})}{\|\mathbf{y}-\mathbf{p}^*(\mathbf{y})\|_2},
\quad \|\bm{\nabla}\mathcal{D}(\mathbf{y})\|_2 = 1
\end{equation}

In the precomputed discrete grid, we estimate the gradient
$\bm{\nabla}\mathcal{D}$ and the Hessian
$\mathbf{H}_{\mathcal{D}}\in\mathbb{R}^{2\times2}$ by central finite differences
over neighboring cells, with one-sided differences at grid borders.
We compute the Hessian as the centered difference of the normalized
gradients $\hat{\bm{\nabla}}\mathcal{D}$.
Unlike the analytical expression in \eqref{eq:dmap:hessian},
this approximation does not guarantee symmetry, motivating our representation
level that stores all four entries.
\begin{equation}
\label{eq:dmap:hessian}
\mathbf{H}_{\mathcal{D}}(\mathbf{y}) =
\begin{bsmallmatrix}
\frac{(y_2 - p_y^*(\mathbf{y}))^2}{\|\mathbf{y} - \mathbf{p}^*(\mathbf{y})\|_2^3} &
-\frac{(y_1 - p_x^*(\mathbf{y}))(y_2 - p_y^*(\mathbf{y}))}{\|\mathbf{y} - \mathbf{p}^*(\mathbf{y})\|_2^3}\\
-\frac{(y_1 - p_x^*(\mathbf{y}))(y_2 - p_y^*(\mathbf{y}))}{\|\mathbf{y} - \mathbf{p}^*(\mathbf{y})\|_2^3} &
\frac{(y_1 - p_x^*(\mathbf{y}))^2}{\|\mathbf{y} - \mathbf{p}^*(\mathbf{y})\|_2^3}
\end{bsmallmatrix}
\end{equation}

\subsection{Representation levels}\label{sec:dmap:levels}

Each cell may store different amounts of precomputed information. We consider
for this work three representation levels $\mathcal{M}_l$ of increasing
information and memory cost, which determine the quantities available at $O(1)$
query cost.
\begin{itemize}
\item \textbf{Level~0 ($\mathcal{M}_0$):} each cell stores only the integer
      index $k^*(\mathbf{y})$ of the nearest reference point in $\mathcal{P}$.
      The distance, gradient, and Hessian are recovered on demand from their
      analytical expressions (see \eqref{eq:dmap:def}, \eqref{eq:dmap:grad} and
      \eqref{eq:dmap:hessian}). $\mathcal{M}_0$ is the most memory-efficient
      layout and serves as an $O(1)$ drop-in replacement for the
      kd-tree~\cite{bentley:cacm:1975} search of ICP, reducing
      the per-query cost from $O(\log n)$ to $O(1)$;
\item \textbf{Level~1 ($\mathcal{M}_1$):} each cell stores the scalar distance
      $\mathcal{D}$ together with the two components of the gradient
      $\bm{\nabla}\mathcal{D}$ (optionally also $k^*$). The Hessian, if needed,
      is recovered at query time through centered difference approximation;
\item \textbf{Level~2 ($\mathcal{M}_2$):} each cell additionally stores all
      entries of the finite-difference Hessian
      $\mathbf{H}_{\mathcal{D}}$, ensuring the second-order information is
      available at $O(1)$.
\end{itemize}

\subsection{Data structures}\label{sec:dmap:datastruct}

Independently of the representation level, we build the three distance map
representations considered in this work by a best-first
wavefront propagation over an 8-connected neighborhood, analogous to Dijkstra's
shortest-path algorithm~\cite{dijkstra:nm:1959}: the cells occupied by reference
points are seeded at zero distance and the nearest-neighbor index is propagated
outward, either over the whole map or until a truncation distance $d_{\max}$ is
reached. Then, post-processing passes compute the gradient and Hessian by finite
differences.

We implement the distance map representations with three data structures,
summarized in \tabref{tab:distance-maps:alignment:dmaps}:
\emph{Nano} ($\mathcal{M}_{0,\mathrm{dense}}$), a dense
integer-index array with the lowest memory footprint; \emph{Standard}
($\mathcal{M}_{1|2,\mathrm{dense}}$), a dense array of full cells; and
\emph{Sparse} ($\mathcal{M}_{1|2,\mathrm{sparse}}$), a hash table
(\texttt{std::unordered\_map} in C++) that stores only the cells within the
truncation band $\mathcal{D}\leq d_{\max}$, analogous to the truncated signed
distance fields used in volumetric reconstruction~\cite{curless-levoy:siggraph:1996}.
All three data structures provide $O(1)$ lookup, with the sparse variant
achieving it on average and mitigating hash-collision worst cases through a
post-propagation rehash.

\begin{table}[!t]
\centering
\caption{Comparison of the Nano, Standard, and Sparse\\Distance Map Variants}
\label{tab:distance-maps:alignment:dmaps}
\begin{tabular}{lcccc}
\hline
\textbf{Variant} & \textbf{Storage} & \textbf{Values/cell} &
\textbf{Query} & $\mathbf{d_{\max}}$? \\
\hline
\emph{Nano} ($\mathcal{M}_{0,\mathrm{dense}}$) &
    Dense array &
    $1$ (\texttt{int}) &
    $O(1)$ &
    optional \\
\emph{Standard} ($\mathcal{M}_{1|2,\mathrm{dense}}$) &
    Dense array &
    $3$--$8^{\star}$ &
    $O(1)$ &
    optional \\
\emph{Sparse} ($\mathcal{M}_{1|2,\mathrm{sparse}}$) &
    Hash map &
    $3$--$8^{\star}$ &
    $O(1)^{\dagger}$ &
    advisable \\
\hline
\multicolumn{5}{l}{%
    \footnotesize
    $^{\star}$ Depends on the representation level:
    $3$ values for $\mathcal{M}_1$,
    $7$ for $\mathcal{M}_2$,}\\
\multicolumn{5}{l}{%
    \footnotesize
    \phantom{$^{\star}$} $+1$ if the optional nearest-neighbor index
    $k^*(\mathbf{y})$ is stored.}\\
\multicolumn{5}{l}{%
    \footnotesize
    $^{\dagger}$ average $O(1)$; worst-case $O(N_d)$ under hash
    collision, mitigated by}\\
\multicolumn{5}{l}{%
    \footnotesize
    \phantom{$^{\dagger}$}
    the post-propagation rehash, where $N_d \leq N_G$ represent
    the number}\\
\multicolumn{5}{l}{%
    \footnotesize
    \phantom{$^{\dagger}$}
    of populated cells within $d_{\max}$ and $N_G$
    the total number of cells.}\\
\end{tabular}
\end{table}


\section{Registration Methodology}\label{sec:method}

This section describes the proposed 2D point cloud registration. We
introduce two constraint factors on $SE(2)$ built on top of the unsigned
distance map $\mathcal{D}$ of \secref{sec:dmap}: a point-to-point residual
(\secref{sec:method:p2p}) and a point-to-plane residual
(\secref{sec:method:p2pl}). Both replace the per-iteration nearest-neighbor
search of ICP with $O(1)$ grid lookups of $\mathcal{D}$ and its derivatives.
We implement these constraint factors on the srrg2\_solver
framework~\cite{grisetti:robotics:2020} with the manifold Gauss-Newton
optimizer.

\subsection{Problem formulation}\label{sec:method:problem}

Let $\mathbf{z}_i\in\mathbb{R}^2$, $i=1,\dots,M$, denote the points of the
moving (current) cloud expressed in the sensor frame $S$. The goal of
registration is to find the robot pose $\mathbf{x}\in SE(2)$ that best aligns
the moving cloud to the reference one. The pose is a homogeneous
transformation $\mathbf{x}=[\mathbf{R}\mid\mathbf{t}]$, with
$\mathbf{R}(\theta)\in SO(2)$ and $\mathbf{t}\in\mathbb{R}^2$. Let
$\mathbf{T}_S^R\in SE(2)$ be the known sensor-to-robot transformation,
with inverse $\mathbf{T}_R^S=[\mathbf{R}_R^S\mid\mathbf{t}_R^S]$. A moving point
$\mathbf{z}_i$ projected by the current estimate and expressed in the reference
sensor frame is represented by $\tilde{\mathbf{z}}_i$,
as in \eqref{eq:method:proj}.
\begin{equation}\label{eq:method:proj}
\tilde{\mathbf{z}}_i = \mathbf{T}_R^S\,\mathbf{x}\,\mathbf{T}_S^R\,\mathbf{z}_i,
\qquad
\mathbf{z}_{R,i} = \mathbf{T}_S^R\,\mathbf{z}_i .
\end{equation}

The problem is formulated as a non-linear least-squares problem and solved with
a Gauss-Newton optimizer on the $SE(2)$ manifold. Because the optimization
variable lives on a manifold, increments cannot be applied by Euclidean
addition. A perturbation $\mathbf{\Delta x}=(\Delta x,\Delta y,\Delta\theta)^T
\in\mathbb{R}^3$ is defined in the tangent space and applied through a
right-sided update
$\mathbf{x}_{k+1}=\mathbf{x}_k\oplus\mathbf{\Delta x}=
\mathbf{x}_k\cdot\exp(\mathbf{\Delta x})$, with
$\exp(\mathbf{\Delta x})=[\mathbf{R}(\Delta\theta)\mid(\Delta x,\Delta y)^T]$.
For the details of the manifold least-squares machinery (the Gauss-Newton
normal equations, robust kernels, and information-matrix recovery for
uncertainty estimation), we follow
its formulation and the srrg2\_solver implementation of
Grisetti~\etal~\cite{grisetti:robotics:2020}. We use the 2D skew operator
$\lfloor\mathbf{v}\rfloor=(-v_y,v_x)^T$, which encodes the rotational
sensitivity of a vector and populates the column associated with $\Delta\theta$
in the Jacobians derived below.

\subsection{Point-to-point error formulation}\label{sec:method:p2p}

First, we propose a point-to-point residual as the scalar returned by the
distance map $\mathcal{D}$ at the projected moving point $\tilde{\mathbf{z}}_i$,
as in \eqref{eq:method:p2p:err}.
This residual measures the Euclidean distance from $\tilde{\mathbf{z}}_i$ to its
nearest reference point, as encoded by the precomputed field $\mathcal{D}$.
\begin{equation}\label{eq:method:p2p:err}
e_i^{\mathrm{p2p}}(\mathbf{x}) =
\mathcal{D}\!\left(\tilde{\mathbf{z}}_i\right) =
\mathcal{D}\!\left(
  \mathbf{T}_R^S \cdot \mathbf{x} \cdot \mathbf{T}_S^R \mathbf{z}_i
\right) \in \mathbb{R}
\end{equation}

Applying the right-sided perturbation and differentiating \wrt
$\mathbf{\Delta x}$ at $\mathbf{\Delta x}=\mathbf{0}$ through the chain rule
yields the $1\times3$ Jacobian of \eqref{eq:method:p2p:jac}.
The first factor is the distance-map gradient at the projected point.
The second one, $\mathbf{J}_i^{\mathrm{p2p,ICP}}$, coincides with the Jacobian
of the classical point-to-point ICP error~\cite{besl-mckay:tpami:1992}.
The gradient $\bm{\nabla}\mathcal{D}(\tilde{\mathbf{z}}_i)$ is retrieved at
$O(1)$ from the $\mathcal{M}_1$ (or $\mathcal{M}_2$) representation, or
evaluated analytically at $\mathcal{M}_0$.
\begin{equation}\label{eq:method:p2p:jac}
\mathbf{J}_i^{\mathrm{p2p}} =
\bm{\nabla}\mathcal{D}\!\left(\tilde{\mathbf{z}}_i\right)^T \cdot
  \underbrace{
    \mathbf{R}_R^S \mathbf{R}(\theta)
    \cdot \left(\mathbf{I}_{2\times2} \mid \lfloor\mathbf{z}_{R,i}\rfloor\right)
  }_{\mathbf{J}_i^{\mathrm{p2p},\mathrm{ICP}}}
\in \mathbb{R}^{1\times3}
\end{equation}

\subsection{Point-to-plane error formulation}\label{sec:method:p2pl}

Our point-to-plane variant incorporates surface-normal information to improve
convergence in environments dominated by planar structures, analogously to Chen
and Medioni~\cite{chen-medioni:icra:1991}. Normals $\mathbf{n}_{z_i}$ are
estimated on the moving cloud (\secref{sec:method:normals}). Let
$\hat{\bm{\nabla}}\mathcal{D}(\tilde{\mathbf{z}}_i)$ be the normalized gradient.
The proposed residual projects the rotated moving normal onto the vector from
the moving point to its nearest reference point, encoded by
$\hat{\bm{\nabla}}\mathcal{D}\cdot\mathcal{D}$, as in
\eqref{eq:method:p2pl:err}.
\begin{equation}\label{eq:method:p2pl:err}
e_i^{\mathrm{p2pl}}(\mathbf{x}) =
\left(\mathbf{R}(\theta)\,\mathbf{n}_{z_i}\right)^T
\hat{\bm{\nabla}}\mathcal{D}(\tilde{\mathbf{z}}_i)\,
\mathcal{D}(\tilde{\mathbf{z}}_i)
\end{equation}

Differentiating \eqref{eq:method:p2pl:err} \wrt $\mathbf{\Delta x}$ gives the
$1\times3$ Jacobian of \eqref{eq:method:p2pl:jac}. The first term accounts for
the rotation of the normal under $\Delta\theta$. The second and third represent
the change in gradient direction (via the Hessian
$\mathbf{H}_{\mathcal{D}}$) and in distance value,
respectively, both sharing the factor $\mathbf{J}_i^{\mathrm{p2p,ICP}}$.
\begin{equation}\label{eq:method:p2pl:jac}
\begin{aligned}
\mathbf{J}_i^{\mathrm{p2pl}} & =
\mathbf{n}_{z_i}^{T}\!
\left(\mathbf{0}_{2\times2}\mid
\lfloor-\mathbf{R}(\theta)^T\hat{\bm{\nabla}}\mathcal{D}(\tilde{\mathbf{z}}_i)\,
\mathcal{D}(\tilde{\mathbf{z}}_i)\rfloor\right)\\
& +
\left(\mathbf{R}(\theta)\mathbf{n}_{z_i}\right)^T
\mathbf{H}_{\mathcal{D}}(\tilde{\mathbf{z}}_i)\,
\mathcal{D}(\tilde{\mathbf{z}}_i)\,\mathbf{J}_i^{\mathrm{p2p,ICP}}\\
& +
\left(\mathbf{R}(\theta)\mathbf{n}_{z_i}\right)^T
\hat{\bm{\nabla}}\mathcal{D}(\tilde{\mathbf{z}}_i)\,
\bm{\nabla}\mathcal{D}(\tilde{\mathbf{z}}_i)^T\mathbf{J}_i^{\mathrm{p2p,ICP}}
\end{aligned}
\end{equation}

When the $\mathcal{M}_0$ representation is used, the gradient and distance are
evaluated analytically and the residual of \eqref{eq:method:p2pl:err} reduces to
the dot product of the rotated moving normal with the displacement
$(\tilde{\mathbf{z}}_i-\mathbf{p}^*)$, which is exactly the classical
point-to-plane ICP residual~\cite{chen-medioni:icra:1991}.
Thus, the two formulations are analytically identical at $\mathcal{M}_0$,
a property confirmed numerically in \secref{sec:results}.
The second and third terms require the Hessian and gradient,
both retrieved at $O(1)$ from the $\mathcal{M}_2$ representation.

\subsection{Surface-normal estimation}\label{sec:method:normals}

Normals are estimated once per scan, before optimization. Exploiting the
angular ordering of the 2D~laser scan, a spatially coherent neighborhood
$\mathcal{N}_r(\mathbf{z}_i)$ of radius $r$ is collected by traversing the
ordered scan in both directions from $i$ (wrapping at the scan boundaries)
until the Euclidean distance to $\mathbf{z}_i$ exceeds $r$, avoiding a
kd-tree search~\cite{bentley:cacm:1975}. If the neighborhood has at least
$N_{\min}\geq 3$ points, the $2\times2$ covariance $\mathbf{C}_i$ is computed
and the normal $\mathbf{n}_{z_i}$ is taken as the eigenvector of least variance
via PCA. The sign ambiguity is resolved by orienting the
normal toward the sensor origin ($\mathbf{n}_{z_i}^T\mathbf{z}_i\leq 0$), which
reflects that obstacle-boundary normals consistently face the scanner. Points
with insufficient neighbors are marked invalid for the point-to-plane term.


\section{Experimental Results}\label{sec:results}

We evaluate the proposed distance map registration framework on two experiments.
A synthetic benchmark (\secref{sec:results:synthetic}) measures
pairwise registration accuracy and uncertainty estimation against known
ground-truth poses under controlled conditions, whereas a real-world
laser-odometry evaluation (\secref{sec:results:iilabs3d}) assesses odometric
drift over extended trajectories. Both experiments compare the proposed
distance map point-to-point ($\mathcal{D}_\mathrm{p2p}$,
\secref{sec:method:p2p}) and point-to-plane ($\mathcal{D}_\mathrm{p2pl}$,
\secref{sec:method:p2pl}) formulations against the standard
point-to-point~\cite{besl-mckay:tpami:1992} and
point-to-plane~\cite{chen-medioni:icra:1991} ICP error metrics
($\mathrm{ICP}_{\mathcal{D},\,\mathrm{p2p}}$ and
$\mathrm{ICP}_{\mathcal{D},\,\mathrm{p2pl}}$, respectively), whose data
association is solved through distance map representations.
All formulations are solved with a common Gauss-Newton optimizer on the $SE(2)$
manifold within the srrg2\_solver framework~\cite{grisetti:robotics:2020},
ensuring consistent conditions across methods. We evaluate $\mathcal{D}_\mathrm{p2p}$ at
representation levels $\mathcal{M}_0$ and $\mathcal{M}_1$,
$\mathcal{D}_\mathrm{p2pl}$ at $\mathcal{M}_0$ and $\mathcal{M}_2$, and both ICP
baselines at $\mathcal{M}_0$.

\subsection{Synthetic benchmark}
\label{sec:results:synthetic}

\paragraph{Experimental setup}
Synthetic 2D~scans are generated with a laser-scanner range model parametrized
based on the Hokuyo~UST-10LX sensor ($\SI{30}{\metre}$ maximum range,
$\SI{0.25}{\degree}$ angular resolution, and $\SI{0.03}{\metre}$ repeatability
used as the range-noise standard deviation) and with a $\SI{360}{\degree}$
field-of-view. Three planar scenes are considered:
a \emph{circle} ($\SI{10}{\metre}$ radius observed from its interior,
rotationally symmetric, and thus, rotationally degenerate);
a \emph{square} ($\SI{20}{\metre}$ side, constraining all three degrees of
freedom of the $SE(2)$ pose up to symmetry); and a \emph{corridor} (two
parallel $\SI{100}{\metre}$ walls $\SI{10}{\metre}$ apart, where translation
along the corridor axis is unobservable because the $\SI{30}{\metre}$ range does
not reach the endpoints). Each scene is evaluated over $N=100$ independent
trials with the true pose $\mathbf{x}^{*}\in SE(2)$ drawn uniformly
(translations up to $\pm\SI{1.5}{\metre}$, reduced for the corridor scene;
rotations up to $\pm\SI{20}{\degree}$). Zero-mean Gaussian noise with
$\sigma\in\{0,\,\SI{0.03}{\metre}\}$ is added independently to both scans,
modelling the noise-free and Hokuyo~UST-10LX cases.

All methods adopt Gauss-Newton on $SE(2)$ (up to 50~iterations,
step-norm termination threshold $10^{-5}$) and a $d_{\max}=\SI{5}{\metre}$
truncation distance. Surface normals for the point-to-plane
formulations are estimated via PCA over a $\SI{0.15}{\metre}$ neighborhood
(minimum of 3~points). Results are reported with and without point cloud
voxelization at the cell resolution~$\delta$ and with and without bi-linear
sub-pixel interpolation (flags~V and~S in the tables; S~applies only to
$\mathcal{M}_1$ and $\mathcal{M}_2$).
\tabref{tab:distance-maps:results:synthetic:accuracy:res-0.03m}
and~\tabref{tab:distance-maps:results:synthetic:accuracy:res-0.05m} report the
results of the synthetic benchmark for the $\delta=\SI{0.03}{\metre}$ and
$\delta=\SI{0.05}{\metre}$ resolutions.

\begin{table*}[!t]
\centering
\caption[%
    Synthetic Benchmark Results for Point Cloud Alignment using
    Proposed Distance Map-based Error Formulations and Distance Map-guided
    ICP Baselines
    (Cell Resolution $\delta = \SI{0.03}{\metre}$)]{%
    Synthetic Benchmark Results for Point Cloud Alignment using
    Proposed Distance Map-based Error Formulations\\
    and Distance Map-guided
    ICP Baselines~\cite{besl-mckay:tpami:1992,chen-medioni:icra:1991}
    (Cell Resolution $\delta = \SI{0.03}{\metre}$)}
\label{tab:distance-maps:results:synthetic:accuracy:res-0.03m}
\begin{tabular}{
    l     
    l     
    cc |  
    S[table-format=1.3] S[table-format=4.2]                     |   
    S[table-format=1.3] S[table-format=4.2]                     |   
    S[table-format=1.3] S[table-format=1.2] S[table-format=1.2] |   
    S[table-format=1.3] S[table-format=1.2] S[table-format=1.2] |   
    S[table-format=1.3] S[table-format=1.2] S[table-format=3.2] |   
    S[table-format=1.3] S[table-format=1.2] S[table-format=3.2]     
}
\hline
\multicolumn{4}{c|}{} &
\multicolumn{4}{c|}{\textbf{Circle}} &
\multicolumn{6}{c|}{\textbf{Square}} &
\multicolumn{6}{c}{\textbf{Corridor}}\\
\cline{5-20}
\multicolumn{4}{c|}{} &
\multicolumn{2}{c|}{$\mathbf{\sigma=\SI{0.00}{\metre}}$} &
\multicolumn{2}{c|}{$\mathbf{\sigma=\SI{0.03}{\metre}}$} &
\multicolumn{3}{c|}{$\mathbf{\sigma=\SI{0.00}{\metre}}$} &
\multicolumn{3}{c|}{$\mathbf{\sigma=\SI{0.03}{\metre}}$} &
\multicolumn{3}{c|}{$\mathbf{\sigma=\SI{0.00}{\metre}}$} &
\multicolumn{3}{c}{$\mathbf{\sigma=\SI{0.03}{\metre}}$}\\
\multicolumn{4}{c|}{} &
\multicolumn{1}{c}{$\varepsilon_t$} &
\multicolumn{1}{c|}{$\chi_{\boldsymbol{\varepsilon}}^2$} &
\multicolumn{1}{c}{$\varepsilon_t$} &
\multicolumn{1}{c|}{$\chi_{\boldsymbol{\varepsilon}}^2$} &
\multicolumn{1}{c}{$\varepsilon_t$} &
\multicolumn{1}{c}{$\varepsilon_\theta$} &
\multicolumn{1}{c|}{$\chi_{\boldsymbol{\varepsilon}}^2$} &
\multicolumn{1}{c}{$\varepsilon_t$} &
\multicolumn{1}{c}{$\varepsilon_\theta$} &
\multicolumn{1}{c|}{$\chi_{\boldsymbol{\varepsilon}}^2$} &
\multicolumn{1}{c}{$\varepsilon_y$} &
\multicolumn{1}{c}{$\varepsilon_\theta$} &
\multicolumn{1}{c|}{$\chi_{\boldsymbol{\varepsilon}}^2$} &
\multicolumn{1}{c}{$\varepsilon_y$} &
\multicolumn{1}{c}{$\varepsilon_\theta$} &
\multicolumn{1}{c}{$\chi_{\boldsymbol{\varepsilon}}^2$}\\
\multicolumn{1}{l}{\textbf{Meth.}} &
\multicolumn{1}{c}{$\mathcal{M}_l$} &
\multicolumn{1}{c}{\textbf{V?}} &
\multicolumn{1}{c|}{\textbf{S?}} &
\multicolumn{1}{c}{(m)} &
\multicolumn{1}{c|}{} &
\multicolumn{1}{c}{(m)} &
\multicolumn{1}{c|}{} &
\multicolumn{1}{c}{(m)} &
\multicolumn{1}{c}{(\textdegree{})} &
\multicolumn{1}{c|}{} &
\multicolumn{1}{c}{(m)} &
\multicolumn{1}{c}{(\textdegree{})} &
\multicolumn{1}{c|}{} &
\multicolumn{1}{c}{(m)} &
\multicolumn{1}{c}{(\textdegree{})} &
\multicolumn{1}{c|}{} &
\multicolumn{1}{c}{(m)} &
\multicolumn{1}{c}{(\textdegree{})} &
\multicolumn{1}{c}{}\\
\hline


\multirow{6}{*}{$\mathcal{D}_\mathrm{p2p}$} & \multirow{2}{*}{$\mathcal{M}_0$}
  & \xmark & -- &               
   0.001 &           6234.68 &  
\B 0.003 &           3100.86 &  
   0.001 &    0.19 &    0.94 &  
   0.003 &    0.08 &    0.19 &  
   0.007 &    0.11 &  540.06 &  
   0.003 &    0.12 &  453.71 \\ 
& & \cmark & -- &               
   0.009 &           4827.02 &  
   0.010 &           3123.95 &  
   0.012 &    0.01 &    0.10 &  
   0.006 &    0.02 &    0.04 &  
   0.009 &    0.10 &  462.70 &  
   0.003 &    0.11 &  461.08 \\ 
\cline{2-20}
& \multirow{4}{*}{$\mathcal{M}_1$}
  & \xmark & \xmark &           
   0.018 &            586.23 &  
\B 0.003 &            934.92 &  
   0.015 &    0.02 &    0.07 &  
   0.003 & \B 0.01 & \B 0.00 &  
   0.008 &    0.21 &   88.20 &  
   0.005 &    0.20 &  158.98 \\ 
& & \xmark & \cmark &           
   0.029 &            257.09 &  
   0.021 &            536.29 &  
   0.051 &    0.02 &    0.32 &  
   0.021 & \B 0.01 &    0.12 &  
   0.011 &    0.22 &   70.90 &  
   0.023 &    0.21 &  113.04 \\ 
& & \cmark & \xmark &           
   0.018 &            584.83 &  
\B 0.003 &            937.83 &  
   0.015 &    0.02 &    0.07 &  
   0.003 & \B 0.01 & \B 0.00 &  
   0.008 &    0.20 &   98.33 &  
   0.005 &    0.20 &  158.77 \\ 
& & \cmark & \cmark &           
   0.029 &            257.08 &  
   0.021 &            536.49 &  
   0.051 &    0.02 &    0.32 &  
   0.021 & \B 0.01 &    0.12 &  
   0.011 &    0.22 &   75.83 &  
   0.024 &    0.20 &  112.46 \\ 

\hline


\multirow{6}{*}{$\mathcal{D}_\mathrm{p2pl}$} & \multirow{2}{*}{$\mathcal{M}_0$}
  & \xmark & -- &                 
\B 0.000 &           \B 52.86 &   
\B 0.003 &          \B 189.38 &   
\B 0.000 & \B 0.00 &  \B 0.00 &   
\B 0.002 &    0.02 &     0.01 &   
\B 0.000 & \B 0.00 &     5.72 &   
\B 0.001 &    0.02 &    13.83 \\  
& & \cmark & -- &                 
\B 0.000 &              54.10 &   
   0.004 &             190.31 &   
\B 0.000 & \B 0.00 &  \B 0.00 &   
\B 0.002 & \B 0.01 &     0.01 &   
\B 0.000 & \B 0.00 &     5.23 &   
   0.002 & \B 0.01 & \B 12.94 \\  
\cline{2-20}
& \multirow{4}{*}{$\mathcal{M}_2$}
  & \xmark & \xmark &             
   0.009 &             463.03 &   
\B 0.003 &            1716.85 &   
   0.004 &    0.03 &     0.01 &   
   0.003 &    0.02 &     0.01 &   
   0.009 &    0.01 &  \B 4.93 &   
\B 0.001 & \B 0.01 &   129.01 \\  
& & \xmark & \cmark &             
   0.023 &             264.87 &   
   0.021 &            1120.07 &   
   0.023 &    0.08 &     0.33 &   
   0.021 & \B 0.01 &     0.30 &   
   0.012 &    0.01 &     8.66 &   
   0.015 & \B 0.01 &    83.51 \\  
& & \cmark & \xmark &             
   0.008 &             462.02 &   
\B 0.003 &            1716.63 &   
   0.004 &    0.03 &     0.01 &   
   0.003 &    0.02 &     0.01 &   
   0.009 &    0.02 &     5.09 &   
\B 0.001 & \B 0.01 &   126.92 \\  
& & \cmark & \cmark &             
   0.023 &             265.03 &   
   0.021 &            1119.63 &   
   0.024 &    0.07 &     0.36 &   
   0.021 & \B 0.01 &     0.30 &   
   0.012 &    0.01 &    11.40 &   
   0.015 & \B 0.01 &    82.42 \\  

\hline


\multirow{2}{*}{$\mathrm{ICP}_{\mathcal{D},\,\mathrm{p2p}}$~\cite{besl-mckay:tpami:1992}}
& \multirow{2}{*}{$\mathcal{M}_0$}
  & \xmark & -- &                 
\B 0.000 &           6432.74 &    
\B 0.003 &           6479.07 &    
   0.001 &    0.19 &    2.07 &    
   0.003 &    0.08 &    0.39 &    
\B 0.000 &    0.01 &  834.29 &    
\B 0.001 &    0.02 &  854.64 \\   
& & \cmark & -- &                 
   0.009 &           6578.50 &    
   0.010 &           6579.53 &    
   0.013 &    0.01 &    0.25 &    
   0.006 &    0.02 &    0.08 &    
\B 0.000 &    0.01 &  775.70 &    
\B 0.001 & \B 0.01 &  817.46 \\   

\hline


\multirow{2}{*}{$\mathrm{ICP}_{\mathcal{D},\,\mathrm{p2pl}}$~\cite{chen-medioni:icra:1991}}
& \multirow{2}{*}{$\mathcal{M}_0$}
  & \xmark & -- &                 
\B 0.000 &           \B 52.86 &   
\B 0.003 &          \B 189.38 &   
\B 0.000 & \B 0.00 &  \B 0.00 &   
\B 0.002 &    0.02 &     0.01 &   
\B 0.000 & \B 0.00 &     5.72 &   
\B 0.001 &    0.02 &    13.83 \\  
& & \cmark & -- &                 
\B 0.000 &              54.10 &   
   0.004 &             190.31 &   
\B 0.000 & \B 0.00 &  \B 0.00 &   
\B 0.002 & \B 0.01 &     0.01 &   
\B 0.000 & \B 0.00 &     5.23 &   
   0.002 & \B 0.01 & \B 12.94 \\  

\hline

\multicolumn{20}{l}{%
    \footnotesize
    Legend:
    distance map representation level~($\mathcal{M}_l$);
    point cloud voxelization~(V);
    sub-pixelization~(S) (bi-linear interpolation); enabled~(\cmark),}\\
\multicolumn{20}{l}{%
    \footnotesize
    disabled~(\xmark), not applicable~(--);
    translation error $\varepsilon_t$~(m);
    rotation error $\varepsilon_\theta$~(\textdegree{});
    pose-error consistency metric
    $\chi_{\boldsymbol{\varepsilon}}^2 = \boldsymbol{\varepsilon}^T\mathbf{\Omega}\,\boldsymbol{\varepsilon}$~(dimensionless).}\\

\end{tabular}
\end{table*}

\begin{table*}[!t]
\centering
\caption[%
    Synthetic Benchmark Results for Point Cloud Alignment using
    Proposed Distance Map-based Error Formulations and Distance Map-guided
    ICP Baselines
    (Cell Resolution $\delta = \SI{0.05}{\metre}$)]{%
    Synthetic Benchmark Results for Point Cloud Alignment using
    Proposed Distance Map-based Error Formulations\\
    and Distance Map-guided
    ICP Baselines~\cite{besl-mckay:tpami:1992,chen-medioni:icra:1991}
    (Cell Resolution $\delta = \SI{0.05}{\metre}$)}
\label{tab:distance-maps:results:synthetic:accuracy:res-0.05m}
\begin{tabular}{
    l     
    l     
    cc |  
    S[table-format=1.3] S[table-format=4.2]                     |   
    S[table-format=1.3] S[table-format=4.2]                     |   
    S[table-format=1.3] S[table-format=1.2] S[table-format=1.2] |   
    S[table-format=1.3] S[table-format=1.2] S[table-format=1.2] |   
    S[table-format=1.3] S[table-format=1.2] S[table-format=3.2] |   
    S[table-format=1.3] S[table-format=1.2] S[table-format=3.2]     
}
\hline
\multicolumn{4}{c|}{} &
\multicolumn{4}{c|}{\textbf{Circle}} &
\multicolumn{6}{c|}{\textbf{Square}} &
\multicolumn{6}{c}{\textbf{Corridor}}\\
\cline{5-20}
\multicolumn{4}{c|}{} &
\multicolumn{2}{c|}{$\mathbf{\sigma=\SI{0.00}{\metre}}$} &
\multicolumn{2}{c|}{$\mathbf{\sigma=\SI{0.03}{\metre}}$} &
\multicolumn{3}{c|}{$\mathbf{\sigma=\SI{0.00}{\metre}}$} &
\multicolumn{3}{c|}{$\mathbf{\sigma=\SI{0.03}{\metre}}$} &
\multicolumn{3}{c|}{$\mathbf{\sigma=\SI{0.00}{\metre}}$} &
\multicolumn{3}{c}{$\mathbf{\sigma=\SI{0.03}{\metre}}$}\\
\multicolumn{4}{c|}{} &
\multicolumn{1}{c}{$\varepsilon_t$} &
\multicolumn{1}{c|}{$\chi_{\boldsymbol{\varepsilon}}^2$} &
\multicolumn{1}{c}{$\varepsilon_t$} &
\multicolumn{1}{c|}{$\chi_{\boldsymbol{\varepsilon}}^2$} &
\multicolumn{1}{c}{$\varepsilon_t$} &
\multicolumn{1}{c}{$\varepsilon_\theta$} &
\multicolumn{1}{c|}{$\chi_{\boldsymbol{\varepsilon}}^2$} &
\multicolumn{1}{c}{$\varepsilon_t$} &
\multicolumn{1}{c}{$\varepsilon_\theta$} &
\multicolumn{1}{c|}{$\chi_{\boldsymbol{\varepsilon}}^2$} &
\multicolumn{1}{c}{$\varepsilon_y$} &
\multicolumn{1}{c}{$\varepsilon_\theta$} &
\multicolumn{1}{c|}{$\chi_{\boldsymbol{\varepsilon}}^2$} &
\multicolumn{1}{c}{$\varepsilon_y$} &
\multicolumn{1}{c}{$\varepsilon_\theta$} &
\multicolumn{1}{c}{$\chi_{\boldsymbol{\varepsilon}}^2$}\\
\multicolumn{1}{l}{\textbf{Meth.}} &
\multicolumn{1}{c}{$\mathcal{M}_l$} &
\multicolumn{1}{c}{\textbf{V?}} &
\multicolumn{1}{c|}{\textbf{S?}} &
\multicolumn{1}{c}{(m)} &
\multicolumn{1}{c|}{} &
\multicolumn{1}{c}{(m)} &
\multicolumn{1}{c|}{} &
\multicolumn{1}{c}{(m)} &
\multicolumn{1}{c}{(\textdegree{})} &
\multicolumn{1}{c|}{} &
\multicolumn{1}{c}{(m)} &
\multicolumn{1}{c}{(\textdegree{})} &
\multicolumn{1}{c|}{} &
\multicolumn{1}{c}{(m)} &
\multicolumn{1}{c}{(\textdegree{})} &
\multicolumn{1}{c|}{} &
\multicolumn{1}{c}{(m)} &
\multicolumn{1}{c}{(\textdegree{})} &
\multicolumn{1}{c}{}\\
\hline


\multirow{6}{*}{$\mathcal{D}_\mathrm{p2p}$} & \multirow{2}{*}{$\mathcal{M}_0$}
  & \xmark & -- &                
   0.001 &           6198.44 &   
\B 0.003 &           3913.39 &   
   0.001 &    0.10 &    0.35 &   
   0.003 &    0.29 &    2.15 &   
   0.018 &    0.07 &  454.93 &   
   0.003 &    0.14 &  492.12 \\  
& & \cmark & -- &                
   0.004 &           4953.51 &   
   0.026 &           3200.64 &   
   0.005 &    0.01 &    0.02 &   
   0.019 &    0.03 &    0.29 &   
   0.013 &    0.11 &  379.95 &   
   0.004 &    0.13 &  522.25 \\  
\cline{2-20}
& \multirow{4}{*}{$\mathcal{M}_1$}
  & \xmark & \xmark &            
   0.022 &            480.88 &   
\B 0.003 &            703.70 &   
   0.027 &    0.05 &    0.03 &   
   0.004 &    0.02 & \B 0.01 &   
   0.019 &    0.22 &   69.63 &   
   0.008 &    0.22 &  110.43 \\  
& & \xmark & \cmark &            
   0.044 &            186.61 &   
   0.039 &            306.68 &   
   0.028 &    0.17 &    0.16 &   
   0.047 & \B 0.01 &    0.48 &   
   0.009 &    0.24 &   44.35 &   
   0.036 &    0.23 &   71.13 \\  
& & \cmark & \xmark &            
   0.022 &            422.14 &   
\B 0.003 &            688.86 &   
   0.027 &    0.05 &    0.03 &   
   0.004 & \B 0.01 & \B 0.01 &   
   0.026 &    0.21 &   76.78 &   
   0.011 &    0.22 &  106.45 \\  
& & \cmark & \cmark &            
   0.044 &            162.18 &   
   0.040 &            304.36 &   
   0.027 &    0.16 &    0.16 &   
   0.048 &    0.02 &    0.50 &   
   0.021 &    0.24 &   45.47 &   
   0.039 &    0.21 &   69.34 \\  

\hline


\multirow{6}{*}{$\mathcal{D}_\mathrm{p2pl}$} & \multirow{2}{*}{$\mathcal{M}_0$}
  & \xmark & -- &                
\B 0.000 &             51.30 &   
\B 0.003 &            183.62 &   
\B 0.000 & \B 0.00 & \B 0.00 &   
\B 0.002 &    0.02 & \B 0.01 &   
\B 0.000 & \B 0.00 &    5.72 &   
\B 0.001 &    0.02 &   13.80 \\  
& & \cmark & -- &                
\B 0.000 &          \B 46.18 &   
   0.007 &         \B 179.93 &   
\B 0.000 & \B 0.00 & \B 0.00 &   
   0.004 & \B 0.01 & \B 0.01 &   
\B 0.000 & \B 0.00 & \B 4.08 &   
   0.003 & \B 0.01 & \B 12.01 \\ 
\cline{2-20}
& \multirow{4}{*}{$\mathcal{M}_2$}
  & \xmark & \xmark &            
   0.017 &            261.79 &   
\B 0.003 &            846.01 &   
   0.030 &    0.06 &    0.08 &   
\B 0.002 & \B 0.01 & \B 0.01 &   
   0.016 &    0.05 &   20.86 &   
\B 0.001 &    0.02 &   56.29 \\  
& & \xmark & \cmark &            
   0.038 &            120.63 &   
   0.036 &            445.88 &   
   0.006 &    0.12 &    0.10 &   
   0.036 & \B 0.01 &    0.58 &   
   0.006 &    0.09 &   15.29 &   
   0.026 &    0.04 &   33.44 \\  
& & \cmark & \xmark &            
   0.017 &            235.66 &   
\B 0.003 &            824.85 &   
   0.030 &    0.06 &    0.09 &   
\B 0.002 & \B 0.01 & \B 0.01 &   
   0.018 &    0.06 &   17.17 &   
\B 0.001 &    0.02 &   53.47 \\  
& & \cmark & \cmark &            
   0.037 &            106.47 &   
   0.036 &            441.65 &   
   0.007 &    0.12 &    0.10 &   
   0.035 & \B 0.01 &    0.57 &   
   0.007 &    0.10 &   12.93 &   
   0.025 &    0.05 &   31.53 \\  

\hline


\multirow{2}{*}{$\mathrm{ICP}_{\mathcal{D},\,\mathrm{p2p}}$~\cite{besl-mckay:tpami:1992}}
& \multirow{2}{*}{$\mathcal{M}_0$}
  & \xmark & -- &                
\B 0.000 &           6431.42 &   
\B 0.003 &           6541.08 &   
   0.001 &    0.10 &    0.61 &   
   0.003 &    0.29 &    4.60 &   
\B 0.000 &    0.03 &  828.40 &   
\B 0.001 &    0.04 &  844.22 \\  
& & \cmark & -- &                
   0.005 &           5754.85 &   
   0.026 &           6248.71 &   
   0.005 &    0.01 &    0.04 &   
   0.020 &    0.03 &    0.58 &   
\B 0.000 &    0.01 &  635.54 &   
   0.002 & \B 0.01 &  766.20 \\  

\hline


\multirow{2}{*}{$\mathrm{ICP}_{\mathcal{D},\,\mathrm{p2pl}}$~\cite{chen-medioni:icra:1991}}
& \multirow{2}{*}{$\mathcal{M}_0$}
  & \xmark & -- &                
\B 0.000 &             51.30 &   
\B 0.003 &            183.62 &   
\B 0.000 & \B 0.00 & \B 0.00 &   
\B 0.002 &    0.02 & \B 0.01 &   
\B 0.000 & \B 0.00 &    5.72 &   
\B 0.001 &    0.02 &   13.80 \\  
& & \cmark & -- &                
\B 0.000 &          \B 46.18 &   
   0.007 &         \B 179.93 &   
\B 0.000 & \B 0.00 & \B 0.00 &   
   0.004 & \B 0.01 & \B 0.01 &   
\B 0.000 & \B 0.00 & \B 4.08 &   
   0.003 & \B 0.01 & \B 12.01 \\ 

\hline

\multicolumn{20}{l}{%
    \footnotesize
    Legend:
    distance map representation level~($\mathcal{M}_l$);
    point cloud voxelization~(V);
    sub-pixelization~(S) (bi-linear interpolation); enabled~(\cmark),}\\
\multicolumn{20}{l}{%
    \footnotesize
    disabled~(\xmark), not applicable~(--);
    translation error $\varepsilon_t$~(m);
    rotation error $\varepsilon_\theta$~(\textdegree{});
    pose-error consistency metric
    $\chi_{\boldsymbol{\varepsilon}}^2 = \boldsymbol{\varepsilon}^T\mathbf{\Omega}\,\boldsymbol{\varepsilon}$~(dimensionless).}\\

\end{tabular}
\end{table*}

\paragraph{Evaluation metrics}
Given an estimate $\mathbf{x}$ and its ground-truth $\mathbf{x}^{*}$, the
target-frame residual transformation
$\mathbf{x}\cdot{\mathbf{x}^{*}}^{-1}=[\mathbf{R}_{\boldsymbol{\varepsilon}}\mid
\mathbf{t}_{\boldsymbol{\varepsilon}}]$ yields the pose error
$\boldsymbol{\varepsilon}(\mathbf{x})$
of~\eqref{eq:results:pose-error}~\cite{kummerle:ar:2009}, from which we derive
the scalar metrics in~\eqref{eq:results:scalar-metrics}.
\begin{equation}\label{eq:results:pose-error}
\mathbf{x}\cdot{\mathbf{x}^{*}}^{-1} =
\left[\mathbf{R}_{\boldsymbol{\varepsilon}}\mid
\mathbf{t}_{\boldsymbol{\varepsilon}}\right],
\;
\boldsymbol{\varepsilon}(\mathbf{x}) =
\left(\mathbf{t}_{\boldsymbol{\varepsilon}}^{T},\,
\theta_{\boldsymbol{\varepsilon}}\right)^{T}\in\mathbb{R}^{3}
\end{equation}
\begin{equation}\label{eq:results:scalar-metrics}
\varepsilon_t = \|\mathbf{t}_{\boldsymbol{\varepsilon}}\|_2,
\,
\varepsilon_y = |t_{\boldsymbol{\varepsilon},y}|,
\,
\varepsilon_\theta =
\left|\atan2\!\left(r_{\boldsymbol{\varepsilon},2,1},
r_{\boldsymbol{\varepsilon},1,1}\right)\right|
\end{equation}

The translation error~$\varepsilon_t$ is reported for the circle and square
scenes. For the corridor, only the observable perpendicular
component~$\varepsilon_y$ is used, since the corridor axis is unobservable. The
rotation error~$\varepsilon_\theta$ is omitted for the circle, whose rotational
DoF is degenerate. The internal consistency of each solver's uncertainty
estimate is assessed by the Hessian-weighted residual norm
of~\eqref{eq:results:chi2}, where $\boldsymbol{\Omega}\in\mathbb{R}^{3\times3}$
is the information matrix returned by the Gauss-Newton
solver~\cite{kummerle:ar:2009,grisetti:robotics:2020}.
\begin{equation}\label{eq:results:chi2}
\chi_{\boldsymbol{\varepsilon}}^2 =
\boldsymbol{\varepsilon}(\mathbf{x})^T\,\boldsymbol{\Omega}\,
\boldsymbol{\varepsilon}(\mathbf{x})
\end{equation}

A value $\chi_{\boldsymbol{\varepsilon}}^2=0$ denotes a perfect alignment, while
large values indicate poor accuracy or an overconfident $\boldsymbol{\Omega}$.
The metric inflates whenever the alignment fails along a degenerate direction,
making it a relative cross-formulation assessment of uncertainty calibration.
The errors $\varepsilon_t$, $\varepsilon_y$, and $\varepsilon_\theta$ are
aggregated as the Root-Mean-Square Error~(RMSE) over the $N=100$ trials, whereas
$\chi_{\boldsymbol{\varepsilon}}^2$ is reported as the mean.

\paragraph{Registration accuracy}
Point-to-plane formulations
($\mathcal{D}_\mathrm{p2pl}$ and $\mathrm{ICP}_{\mathcal{D},\,\mathrm{p2pl}}$)
achieve lower or equal $\varepsilon_t$ and $\varepsilon_\theta$ than their
point-to-point counterparts, confirming the conclusion of Rusinkiewicz and
Levoy~\cite{rusinkiewicz-levoy:im:2001} for ICP variants and extending it to
distance map-based formulations. This behavior reflects the
sensitivity of point-to-point residuals to the inter-scan sampling discrepancy
inherent to the laser-scanner model when two scans sample the surfaces at
different poses even without noise. In contrast, point-to-plane residuals
measure the orthogonal point-to-surface distance. Thus, point-to-plane variants
seem more invariant to that sampling discrepancy.

Moreover, $\mathcal{D}_\mathrm{p2pl}(\mathcal{M}_0)$ and
$\mathrm{ICP}_{\mathcal{D},\,\mathrm{p2pl}}$ produce
identical results in the synthetic benchmark, confirming
the analytical equivalence derived in \secref{sec:method:p2pl}.
Regarding the distance map parametrization,
$\mathcal{D}_\mathrm{p2p}(\mathcal{M}_1)$ accuracy degrades as $\delta$ grows.
Enabling bi-linear sub-pixel interpolation does not improve, and often worsens,
accuracy: at $\delta=\SI{0.05}{\metre}$ with noise and no voxelization,
$\varepsilon_t$ rises from $\SI{0.003}{\metre}$, $\SI{0.004}{\metre}$, and
$\SI{0.008}{\metre}$ to $\SI{0.039}{\metre}$, $\SI{0.047}{\metre}$, and
$\SI{0.036}{\metre}$ for the circle, square, and corridor scenes, respectively.
This result is a consequence of initializing hit cells on the distance map
computation with zero, biasing the interpolated field, even though facilitates
the precomputation procedure. Voxelization has a negligible effect under these
controlled conditions, but remains recommended for 2D~lasers, whose
radial sampling concentrates measurements near the sensor and would otherwise
bias the optimizer toward close-range observations.

The precomputed methods trade a small accuracy loss compared to
their analytical counterparts for the $O(1)$ lookup. In particular,
$\mathcal{D}_\mathrm{p2pl}(\mathcal{M}_2)$ retains the point-to-plane
advantage over the point-to-point formulations yet does not match the analytical
$\mathcal{D}_\mathrm{p2pl}(\mathcal{M}_0)$ and
$\mathrm{ICP}_{\mathcal{D},\,\mathrm{p2pl}}$ (\eg, on the noise-free square at
$\delta=\SI{0.03}{\metre}$, its
$\varepsilon_\theta=\SI{0.03}{\degree}$ stays lower than
$\SI{0.94}{\degree}$ and $\SI{2.07}{\degree}$ of the analytical point-to-point
formulations, but above the $\SI{0.00}{\degree}$ of the analytical
point-to-plane). Two effects contribute to this gap. First, the grid
quantization and the finite-difference gradient introduce a systematic bias in
the residual direction that scales with $\delta$. Second, the point-to-plane
term at $\mathcal{M}_2$ relies on the second-order derivative, obtained by
finite-differencing the already finite-difference gradient. This successive
differentiation is seems more noise-sensitive than the first-order gradient
of $\mathcal{M}_1$, preventing $\mathcal{M}_2$ from matching the analytical
point-to-plane accuracy. Thus, the analytical $\mathcal{M}_0$ point-to-plane
remains the most accurate configuration in these controlled conditions, while
the precomputed levels offer competitive accuracy performance.

\paragraph{Uncertainty estimation}
The consistency metric $\chi_{\boldsymbol{\varepsilon}}^2$ exposes differences
in the estimated information matrices. The
$\mathcal{D}_\mathrm{p2pl}(\mathcal{M}_0)$ and
$\mathrm{ICP}_{\mathcal{D},\,\mathrm{p2pl}}$ point-to-plane formulations yield
the best-calibrated estimates across all conditions, reaching
$\chi_{\boldsymbol{\varepsilon}}^2\approx 0$ on the noise-free square scene.
In contrast, the analytical point-to-point formulations are overconfident on
degenerate scenes: on the circle, $\mathrm{ICP}_{\mathcal{D},\,\mathrm{p2p}}$
yields $\chi_{\boldsymbol{\varepsilon}}^2$ up to
$\approx\!6.6\times10^{3}$, two to three orders of magnitude above the
point-to-plane counterpart ($\approx\!46$--$190$), as its $2\times3$ Jacobian
accumulates a larger information matrix than the scalar $1\times3$ distance map
residual. The precomputed $\mathcal{D}_\mathrm{p2p}(\mathcal{M}_1)$ variant
yields systematically lower $\chi_{\boldsymbol{\varepsilon}}^2$ than both
$\mathcal{D}_\mathrm{p2p}(\mathcal{M}_0)$ and
$\mathrm{ICP}_{\mathcal{D},\,\mathrm{p2p}}$ on the circle and corridor scenes
(\eg, circle, noise-free, $\delta=\SI{0.03}{\metre}$:
$\approx\!257$--$586$ against $\approx\!4.8$--$6.6\times10^{3}$), suggesting
that the finite-difference gradient and the cell-resolution regularize the
information matrix estimation. Conversely, $\mathcal{D}_\mathrm{p2pl}(\mathcal{M}_2)$ seems
more noise-sensitive than its $\mathcal{M}_0$ counterpart (\eg, on
the noisy circle at $\delta=\SI{0.03}{\metre}$,
$\chi_{\boldsymbol{\varepsilon}}^2\approx\!1120$--$1717$), consistent with the
compounded second-order finite differences noted in the accuracy results.
Hence, the analytical $\mathcal{M}_0$ point-to-plane formulations are preferable
for uncertainty-aware applications.

\subsection{Real-world laser odometry evaluation}
\label{sec:results:iilabs3d}

\paragraph{Experimental setup}
We deploy the four formulations ($\mathcal{D}_\mathrm{p2p}$,
$\mathcal{D}_\mathrm{p2pl}$,
$\mathrm{ICP}_{\mathcal{D},\,\mathrm{p2p}}$~\cite{besl-mckay:tpami:1992}, and
$\mathrm{ICP}_{\mathcal{D},\,\mathrm{p2pl}}$~\cite{chen-medioni:icra:1991}) as a
2D~laser-odometry tracker on
the IILABS~3D~\cite{ribeiro:access:2025} dataset, using only the Hokuyo~UST-10LX
scans, with ground-truth from an OptiTrack motion-capture
system. Four planar sequences are considered: \emph{Nav~A~Diff}
($\SI{275}{\metre}$, differential drive motion), \emph{Nav~A~Omni}
($\SI{112}{\metre}$, holonomic motion), \emph{Loop} ($\SI{232}{\metre}$, where
the robot exits and re-enters the Nav~A space), and
\emph{Slippage} ($\SI{39}{\metre}$, a straight path with induced wheel slippage).
The tracker uses $\delta=\SI{0.03}{\metre}$, $d_{\max}=\SI{0.50}{\metre}$, and a
maximum sensor range $r_{\max}=\SI{30}{\metre}$, with bi-linear interpolation disabled and
voxelization enabled following the synthetic findings. Outliers are handled by a
two-stage Cauchy robustifier (5~iterations coarse stage with $c^2=0.250$,
followed by a fine stage of up to 15~iterations with $c^2=0.005\approx2\sigma$).
The solver stops at step-norm $10^{-5}$ or 20~iterations.
Normals are estimated via PCA over a $\SI{0.15}{\metre}$
neighborhood (minimum of 5~points), and the keyframe is reset whenever the
inlier ratio drops below $\SI{20}{\percent}$. Two keyframe-splitting criteria
are considered: a translation
($\SI{0.50}{\metre}$--$\SI{3.00}{\metre}$) and an inlier-ratio threshold
($\SI{40}{\percent}$--$\SI{90}{\percent}$).

\paragraph{Evaluation metrics}
Drift is assessed with relative metrics computed over
$\Delta d=\SI{10}{\metre}$ trajectory segments~\cite{ribeiro:access:2025} using
the evo framework~\cite{grupp:evo} on the estimated and ground-truth
TUM~\cite{sturm:iros:2012} trajectories. For each segment relative error
$\boldsymbol{\varepsilon}_s$ with translational
$\mathbf{t}_{\boldsymbol{\varepsilon}_s}$ and rotational
$\theta_{\boldsymbol{\varepsilon}_s}$ parts over $N_s$ segments, the Relative
Translational Error~(RTE) and Relative Rotational Error~(RRE) are computed as
in~\eqref{eq:results:rte-rre}, where RTE~\cite{geiger:cvpr:2012} expresses
drift as a percentage of the segment length.
\begin{equation}\label{eq:results:rte-rre}
\mathrm{RTE} =
\frac{100\,\%}{N_s}\sum_{s=1}^{N_s}
\frac{\|\mathbf{t}_{\boldsymbol{\varepsilon}_s}\|_2}{\Delta d},
\qquad
\mathrm{RRE} =
\frac{1}{N_s}\sum_{s=1}^{N_s}
\frac{|\theta_{\boldsymbol{\varepsilon}_s}|}{\Delta d}
\end{equation}

Both RTE and RRE are reported as the mean over all
segments~\cite{ribeiro:access:2025}. These relative
metrics are preferred over the absolute trajectory error to isolate the local
consistency of the laser-odometry front-end independently of global alignment.
\tabref{tab:distance-maps:results:iilabs3d:accuracy:rmax-30.0m} reports the
odometric drift results.

{%
\renewcommand{\arraystretch}{0.88}
\setlength{\tabcolsep}{4pt}
\begin{table*}[!t]
\centering
\caption[%
    Real-world Laser Odometry Drift on the IILABS~3D
    Dataset using Proposed Distance Map-based Formulations
    and Distance Map-guided ICP Baselines
    for the 2D~Point Cloud Registration Problem]{%
    Real-world Laser Odometry Drift on the IILABS~3D~\cite{ribeiro:access:2025}
    Dataset using Proposed Distance Map-based Formulations\\
    and Distance Map-guided ICP Baselines~\cite{besl-mckay:tpami:1992,chen-medioni:icra:1991}
    for the 2D~Point Cloud Registration Problem}
\label{tab:distance-maps:results:iilabs3d:accuracy:rmax-30.0m}
\begin{tabular}{
    l   
    l   
    c | 
    S[table-format=2.2] S[table-format=1.3] | 
    S[table-format=2.2] S[table-format=1.3] | 
    S[table-format=2.2] S[table-format=1.3] | 
    S[table-format=2.2] S[table-format=1.3]   
}
\hline
\multicolumn{3}{c|}{} &
\multicolumn{2}{c|}{\textbf{Nav~A Diff}} &
\multicolumn{2}{c|}{\textbf{Nav~A Omni}} &
\multicolumn{2}{c|}{\textbf{Loop}} &
\multicolumn{2}{c}{\textbf{Slippage}}\\
\cline{4-11}
\multicolumn{3}{c|}{} &
\multicolumn{1}{c}{\textbf{RTE}} &
\multicolumn{1}{c|}{\textbf{RRE}} &
\multicolumn{1}{c}{\textbf{RTE}} &
\multicolumn{1}{c|}{\textbf{RRE}} &
\multicolumn{1}{c}{\textbf{RTE}} &
\multicolumn{1}{c|}{\textbf{RRE}} &
\multicolumn{1}{c}{\textbf{RTE}} &
\multicolumn{1}{c}{\textbf{RRE}}\\
\multicolumn{1}{l}{\textbf{Method}} &
\multicolumn{1}{l}{\textbf{Splitting Criteria}} &
\multicolumn{1}{c|}{\textbf{Threshold}} &
\multicolumn{1}{c}{(\%)} &
\multicolumn{1}{c|}{(\si{\degree\per\metre})} &
\multicolumn{1}{c}{(\%)} &
\multicolumn{1}{c|}{(\si{\degree\per\metre})} &
\multicolumn{1}{c}{(\%)} &
\multicolumn{1}{c|}{(\si{\degree\per\metre})} &
\multicolumn{1}{c}{(\%)} &
\multicolumn{1}{c}{(\si{\degree\per\metre})}\\
\hline


\multirow{12}{*}{$\mathcal{D}_\mathrm{p2p}(\mathcal{M}_0)$} & \multirow{5}{*}{\emph{Translation} (m)}
 & $\SI{0.50}{\metre}$ & 2.91     & 0.182     & 2.64     & 0.134     & 5.80     & 0.310     & 1.28     & 0.059      \\
&& $\SI{1.00}{\metre}$ & 2.05     & 0.161     & 1.94     & 0.108     & 4.35     & 0.236     & 0.96     & 0.074      \\
&& $\SI{1.50}{\metre}$ & 1.76     & 0.146     & 1.63     & 0.098     & \U{3.03} & \U{0.183} & 0.78     & 0.056      \\
&& $\SI{2.00}{\metre}$ & 1.51     & 0.127     & 1.50     & 0.102     & 3.13     & 0.202     & 0.97     & 0.049      \\
&& $\SI{2.50}{\metre}$ & \U{1.35} & 0.126     & \U{1.23} & \U{0.088} & 3.86     & 0.199     & \U{0.68} & \UB{0.033} \\
&& $\SI{3.00}{\metre}$ & 1.47     & \U{0.120} & 1.45     & 0.121     & 3.91     & 0.198     & 0.92     & 0.060      \\
\cline{2-11}
& \multirow{5}{*}{\emph{Inliers Ratio} (\%)}
 & 40\% & \U{1.12} & \U{0.098} & \U{1.09} & 0.090     & \U{2.54} & 0.158     & 0.71     & \UB{0.034} \\
&& 50\% & 1.35     & 0.122     & 1.31     & \U{0.089} & 3.00     & \U{0.144} & 0.64     & 0.052      \\
&& 60\% & 1.58     & 0.129     & 1.39     & 0.090     & 3.12     & 0.175     & \U{0.63} & 0.052      \\
&& 70\% & 1.83     & 0.141     & 1.66     & 0.098     & 4.37     & 0.217     & 0.95     & 0.082      \\
&& 80\% & 2.53     & 0.174     & 2.17     & 0.127     & 5.34     & 0.256     & 1.24     & 0.064      \\
&& 90\% & 5.34     & 0.292     & 3.98     & 0.187     & 9.65     & 0.399     & 2.80     & 0.093      \\

\hline


\multirow{12}{*}{$\mathcal{D}_\mathrm{p2p}(\mathcal{M}_1)$} & \multirow{5}{*}{\emph{Translation} (m)}
 & $\SI{0.50}{\metre}$ & 1.62     & 0.109     & 1.38     & 0.080     & 2.87     & \U{0.140} & 1.47      & 0.088     \\
&& $\SI{1.00}{\metre}$ & 1.12     & 0.096     & 1.01     & \U{0.077} & 3.09     & 0.160     & 0.89      & 0.088     \\
&& $\SI{1.50}{\metre}$ & 1.11     & 0.098     & 0.98     & 0.080     & \U{2.81} & 0.154     & 1.04      & 0.142     \\
&& $\SI{2.00}{\metre}$ & \U{1.06} & \U{0.091} & 0.92     & 0.079     & 2.82     & 0.155     & \UB{0.55} & \U{0.056} \\
&& $\SI{2.50}{\metre}$ & 1.12     & \U{0.091} & \U{0.91} & 0.080     & 3.07     & 0.166     & 0.85      & 0.086     \\
&& $\SI{3.00}{\metre}$ & 1.09     & \U{0.091} & 1.36     & 0.123     & 3.03     & 0.156     & 0.88      & 0.069     \\
\cline{2-11}
& \multirow{5}{*}{\emph{Inliers Ratio} (\%)}
 & 40\% & 1.01     & 0.090      & 0.89      & 0.087      & \U{2.14} & 0.134     & 0.66     & 0.076     \\
&& 50\% & \U{0.95} & \UB{0.078} & \UB{0.81} & \B 0.065   & 2.33     & \U{0.123} & \U{0.64} & \U{0.070} \\
&& 60\% & 1.10     & 0.089      & 0.89      & 0.080      & 2.80     & 0.143     & 0.85     & 0.087     \\
&& 70\% & 1.15     & 0.094      & 0.97      & \UB{0.063} & 3.01     & 0.155     & 0.81     & 0.080     \\
&& 80\% & 1.37     & 0.112      & 1.15      & 0.075      & 3.18     & 0.160     & 0.89     & 0.083     \\
&& 90\% & 1.92     & 0.131      & 1.61      & \B 0.067   & 3.99     & 0.133     & 2.05     & 0.144     \\

\hline


\multirow{12}{*}{$\mathcal{D}_\mathrm{p2pl}(\mathcal{M}_0)$} & \multirow{5}{*}{\emph{Translation} (m)}
 & $\SI{0.50}{\metre}$ & 1.60     & 0.145     & 1.35     & 0.103      & 4.13     & 0.275     & 1.48     & 0.149     \\
&& $\SI{1.00}{\metre}$ & 1.36     & 0.115     & 1.30     & 0.106      & 3.75     & 0.239     & 1.01     & 0.091     \\
&& $\SI{1.50}{\metre}$ & 1.25     & 0.126     & 1.19     & 0.096      & 2.88     & 0.210     & 0.77     & \U{0.077} \\
&& $\SI{2.00}{\metre}$ & 1.16     & 0.115     & \U{0.88} & \UB{0.064} & \U{2.39} & 0.163     & 0.71     & 0.097     \\
&& $\SI{2.50}{\metre}$ & \U{1.08} & \U{0.103} & 1.05     & 0.105      & 2.91     & 0.189     & \U{0.66} & 0.081     \\
&& $\SI{3.00}{\metre}$ & 1.67     & 0.114     & 1.25     & 0.108      & 3.59     & \U{0.140} & 0.75     & 0.090     \\
\cline{2-11}
& \multirow{5}{*}{\emph{Inliers Ratio} (\%)}
 & 40\% & \UB{0.92} & \U{0.094} & 1.09      & 0.104     & \UB{1.41} & \UB{0.107} & 0.84     & 0.073     \\
&& 50\% & 1.06      & 0.104     & \UB{0.84} & 0.086     & 2.05      & 0.139      & 0.72     & 0.084     \\
&& 60\% & 1.10      & 0.111     & \B 0.85   & \U{0.084} & 2.77      & 0.156      & \U{0.68} & \U{0.056} \\
&& 70\% & 1.24      & 0.124     & 1.06      & 0.090     & 3.69      & 0.212      & 0.69     & \U{0.056} \\
&& 80\% & 3.38      & 0.154     & 5.40      & 0.233     & 7.10      & 0.253      & 5.01     & 0.084     \\
&& 90\% & 8.02      & 0.216     & 9.48      & 0.214     & 8.24      & 0.139      & 7.06     & 0.251     \\

\hline


\multirow{12}{*}{$\mathcal{D}_\mathrm{p2pl}(\mathcal{M}_2)$} & \multirow{5}{*}{\emph{Translation} (m)}
 & $\SI{0.50}{\metre}$ & 1.34     & 0.128     & 1.07     & 0.097     & 4.43     & 0.279     & 0.90     & 0.084     \\
&& $\SI{1.00}{\metre}$ & 1.30     & 0.120     & 1.11     & 0.109     & 3.93     & 0.246     & 0.92     & 0.097     \\
&& $\SI{1.50}{\metre}$ & 1.24     & 0.121     & 1.12     & 0.102     & 3.83     & 0.227     & \U{0.78} & \U{0.051} \\
&& $\SI{2.00}{\metre}$ & \U{1.13} & \U{0.103} & 1.05     & 0.103     & 3.21     & 0.188     & 1.00     & 0.053     \\
&& $\SI{2.50}{\metre}$ & \U{1.13} & 0.106     & \U{0.98} & \U{0.093} & 3.50     & 0.189     & 0.81     & 0.060     \\
&& $\SI{3.00}{\metre}$ & 1.22     & 0.123     & 1.24     & 0.118     & \U{2.26} & \U{0.156} & 0.97     & 0.112     \\
\cline{2-11}
& \multirow{5}{*}{\emph{Inliers Ratio} (\%)}
 & 40\% &  1.03     & \U{0.087} &  1.00     & \U{0.085} &  \U{2.09} & \U{0.124} &  \U{0.92} & 0.098     \\
&& 50\% &  \U{1.02} & 0.090     &  \U{0.92} & 0.087     &  3.18     & 0.148     &  1.06     & 0.133     \\
&& 60\% &  1.24     & 0.108     &  1.11     & 0.104     &  3.24     & 0.168     &  0.98     & \U{0.050} \\
&& 70\% &  1.37     & 0.129     &  1.11     & 0.106     &  3.87     & 0.225     &  1.16     & 0.062     \\
&& 80\% &  2.60     & 0.174     &  2.62     & 0.216     &  7.61     & 0.250     &  7.99     & 0.240     \\
&& 90\% & 14.02     & 0.630     & 12.10     & 0.577     & 11.33     & 0.611     & 16.03     & 0.472     \\

\hline


\multirow{12}{*}{$\mathrm{ICP}_{\mathcal{D},\,\mathrm{p2p}}$~\cite{besl-mckay:tpami:1992}}
& \multirow{5}{*}{\emph{Translation} (m)}
 & $\SI{0.50}{\metre}$ & 3.03     & 0.180     & 2.56     & 0.133     & 5.63     & 0.297     & 1.20      & 0.070      \\
&& $\SI{1.00}{\metre}$ & 2.01     & 0.154     & 1.76     & 0.117     & 4.46     & 0.235     & \UB{0.57} & 0.067      \\
&& $\SI{1.50}{\metre}$ & 1.56     & 0.133     & 1.50     & 0.107     & 2.97     & \U{0.180} & 0.70      & 0.052      \\
&& $\SI{2.00}{\metre}$ & 1.40     & 0.123     & 1.19     & 0.100     & \U{2.56} & \U{0.180} & 0.67      & 0.056      \\
&& $\SI{2.50}{\metre}$ & \U{1.24} & \U{0.110} & \U{1.13} & \U{0.082} & 3.55     & 0.181     & 0.67      & 0.045      \\
&& $\SI{3.00}{\metre}$ & 1.33     & 0.113     & 1.59     & 0.095     & 3.60     & 0.186     & 0.67      & \UB{0.038} \\
\cline{2-11}
& \multirow{5}{*}{\emph{Inliers Ratio} (\%)}
 & 40\% & \U{1.10} & \U{0.099} & \U{1.07} & 0.092     &  3.36     & 0.192     & \UB{0.56} & 0.055     \\
&& 50\% & 1.32     & 0.116     & 1.16     & \U{0.082} &  \U{2.95} & \U{0.158} & 0.72      & 0.065     \\
&& 60\% & 1.48     & 0.133     & 1.21     & 0.100     &  3.15     & 0.169     & 0.64      & \U{0.048} \\
&& 70\% & 1.80     & 0.147     & 1.60     & 0.110     &  4.32     & 0.209     & 0.88      & 0.070     \\
&& 80\% & 2.54     & 0.169     & 2.14     & 0.111     &  5.24     & 0.247     & 0.83      & 0.057     \\
&& 90\% & 5.14     & 0.298     & 4.01     & 0.177     & 10.01     & 0.403     & 2.63      & 0.099     \\

\hline


\multirow{12}{*}{$\mathrm{ICP}_{\mathcal{D},\,\mathrm{p2pl}}$~\cite{chen-medioni:icra:1991}}
& \multirow{5}{*}{\emph{Translation} (m)}
 & $\SI{0.50}{\metre}$ & 1.61     & 0.146     & 1.42     & 0.111     & 4.14     & 0.275     & 1.48     & 0.149     \\
&& $\SI{1.00}{\metre}$ & 1.37     & 0.114     & 1.28     & 0.103     & 3.71     & 0.238     & 1.06     & 0.095     \\
&& $\SI{1.50}{\metre}$ & 1.25     & 0.122     & 1.19     & 0.097     & 2.82     & 0.207     & 0.77     & \U{0.077} \\
&& $\SI{2.00}{\metre}$ & 1.13     & 0.115     & \U{0.91} & \U{0.070} & \U{2.45} & 0.164     & 0.71     & 0.097     \\
&& $\SI{2.50}{\metre}$ & \U{1.09} & \U{0.095} & 1.02     & 0.105     & 3.02     & 0.187     & \U{0.66} & 0.081     \\
&& $\SI{3.00}{\metre}$ & 1.68     & 0.112     & 1.22     & 0.103     & 3.55     & \U{0.140} & 0.75     & 0.090     \\
\cline{2-11}
& \multirow{5}{*}{\emph{Inliers Ratio} (\%)}
 & 40\% & \UB{0.88} & \UB{0.082} & 1.06      & 0.106     & \UB{1.37} & \UB{0.106} & 0.78     & 0.073     \\
&& 50\% & 1.01      & 0.097      & \B 0.83   & 0.093     & 2.07      & 0.140      & 0.72     & 0.084     \\
&& 60\% & 1.08      & 0.109      & \UB{0.82} & \U{0.075} & 2.71      & 0.154      & 0.82     & 0.070     \\
&& 70\% & 1.24      & 0.123      & 0.93      & 0.090     & 3.91      & 0.207      & \U{0.69} & \U{0.056} \\
&& 80\% & 3.34      & 0.155      & 5.39      & 0.231     & 7.32      & 0.257      & 4.92     & 0.081     \\
&& 90\% & 7.98      & 0.214      & 9.45      & 0.217     & 8.26      & 0.136      & 6.97     & 0.251     \\

\hline

\multicolumn{11}{l}{%
    \footnotesize
    Legend:
    distance map representation level~($\mathcal{M}_l$);
    Relative Translational Error (RTE); Relative Rotational Error (RRE);}\\
\multicolumn{11}{l}{%
    \footnotesize
    RTE (\%) and RRE (\si{\degree\per\metre}) reported as mean over
    \SI{10}{\metre} segments.
    Bold entries indicate lowest drift within \SI{0.05}{\percent} RTE}\\
\multicolumn{11}{l}{%
    \footnotesize
    and \SI{0.005}{\degree\per\metre} RRE.
    Underlined entries indicate lowest drift within method--splitting setups.}\\

\end{tabular}
\end{table*}}

\paragraph{Laser odometry drift}
As in the synthetic benchmark, $\mathcal{D}_\mathrm{p2pl}(\mathcal{M}_0)$ and
$\mathrm{ICP}_{\mathcal{D},\,\mathrm{p2pl}}$ produce numerically equivalent
trajectories (34 and 38~of the 48~configurations within $\SI{0.05}{\percent}$
RTE and $\SI{0.005}{\degree\per\metre}$ RRE, respectively), reaffirming the
theoretical equivalence under real tracking conditions. At the $\mathcal{M}_0$
level, the point-to-plane formulations outperform their point-to-point
counterparts across the majority of configurations, especially on RTE,
consistent with the synthetic results and with
Rusinkiewicz and Levoy~\cite{rusinkiewicz-levoy:im:2001}. The main exception
arises at the $\SI{90}{\percent}$ inlier threshold, where point-to-plane
degrades on Nav~A~Diff, Nav~A~Omni, and Slippage for both the distance map and
ICP variants, indicating that the degradation stems from the error formulation
rather than the correspondence mechanism. The point-to-plane advantage persists
on the Loop sequence, whose doorway transitions keep point-to-point residuals
limited by the sampling discrepancy.

The precomputed point-to-point formulation
$\mathcal{D}_\mathrm{p2p}(\mathcal{M}_1)$ proposed in this work outperforms both
$\mathcal{D}_\mathrm{p2p}(\mathcal{M}_0)$ and
$\mathrm{ICP}_{\mathcal{D},\,\mathrm{p2p}}$ across most configurations
(e.g., lower RTE in 39 and lower RRE in 35~of the 48~cases against
$\mathrm{ICP}_{\mathcal{D},\,\mathrm{p2p}}$), and even reaches competitive drift
against the point-to-plane formulations on Nav~A~Diff, Nav~A~Omni, and Slippage,
with best figures of $\SI{0.95}{\percent}$/$\SI{0.078}{\degree\per\metre}$,
$\SI{0.81}{\percent}$/$\SI{0.063}{\degree\per\metre}$, and
$\SI{0.55}{\percent}$/$\SI{0.056}{\degree\per\metre}$, respectively. This result
is attributed to the finite-difference gradient acting as an implicit
point-to-plane constraint within planar regions while regularizing sensor noise,
without requiring explicit normal estimation. The residual gap persists on the
Loop sequence ($\SI{2.14}{\percent}$ against $\SI{1.37}{\percent}$ for
$\mathrm{ICP}_{\mathcal{D},\,\mathrm{p2pl}}$), where gradient directions deviate
from surface normals near the doorway discontinuities. The precomputed
point-to-plane variant $\mathcal{D}_\mathrm{p2pl}(\mathcal{M}_2)$ is competitive
under the translation-based criterion but consistently worse at the
$\SI{80}{\percent}$--$\SI{90}{\percent}$ inlier thresholds, mirroring its
noise sensitivity observed in the synthetic benchmark.

Finally, the keyframe-splitting criterion has an effect on drift. The
translation-based criterion performs best between $\SI{2.0}{\metre}$ and
$\SI{2.5}{\metre}$. The inlier-ratio criterion performs best at
$\SI{40}{\percent}$--$\SI{60}{\percent}$, degrading beyond
$\SI{80}{\percent}$, where frequent keyframe splits accumulate registration and
numerical errors. In most configurations, the $\SI{40}{\percent}$ inlier
threshold yields lower or equal drift than the best translation threshold,
indicating that a data-driven splitting policy adapts to varying tracking
conditions better than a fixed distance criterion.


\section{Conclusions}\label{sec:conclusions}

This paper proposed a 2D~point cloud registration framework based on unsigned
distance maps with $O(1)$ grid lookups over a precomputed field. We derived
point-to-point and
point-to-plane residuals on the $SE(2)$ manifold. On both the synthetic
benchmark and the real-world IILABS~3D~\cite{ribeiro:access:2025} dataset, the
precomputed point-to-point variant $\mathcal{D}_\mathrm{p2p}(\mathcal{M}_1)$
consistently outperformed its analytical point-to-point counterparts.
Also, $\mathcal{D}_\mathrm{p2p}(\mathcal{M}_1)$ reached
competitive laser-odometry drift even against point-to-plane formulations,
as the finite-difference gradient acts as an implicit point-to-plane constraint
within planar regions while regularizing sensor noise. The uncertainty
calibration analysis showed the analytical $\mathcal{M}_0$
point-to-plane methods to be the best-calibrated on degenerate scenes and
remaining preferable for uncertainty-aware applications. Still,
the precomputed gradient in $\mathcal{D}_\mathrm{p2p}(\mathcal{M}_1)$
regularizes the point-to-point information-matrix
estimate relative to its analytical counterparts.
Future work will assess the computational performance of the proposed framework
against kd-tree~\cite{bentley:cacm:1975}-based ICP,
and integrate it into a full SLAM pipeline.


\bibliographystyle{ieeetr}
\bibliography{IEEEabrv,myrefs}

\end{document}